\documentclass{article}

\usepackage{amsmath,amsfonts,bm}

\def\eqref#1{equation~\ref{#1}}

\def\1{\bm{1}}

\DeclareMathAlphabet{\mathsfit}{\encodingdefault}{\sfdefault}{m}{sl}
\SetMathAlphabet{\mathsfit}{bold}{\encodingdefault}{\sfdefault}{bx}{n}

\DeclareMathOperator*{\argmax}{arg\,max}

\newcommand{\mytitle}{Distributed In-Context Learning \\ under Non-IID Among Clients}

\usepackage{microtype}
\usepackage{booktabs} 
\usepackage{bbm}
\usepackage{mathtools}
\usepackage{nccmath}
\usepackage{setspace}

\usepackage{subcaption}
\usepackage{stackengine,scalerel}
\usepackage{accents}
\usepackage{algorithm} 
\usepackage{algorithmic}  
\usepackage[linesnumbered,ruled,vlined,algo2e]{algorithm2e}
\SetKwInput{KwInput}{Input}                
\SetKwInput{KwOutput}{Output}              

\SetCommentSty{mycommfont}

\SetAlCapSty{algcapsty}

\usepackage[T1]{fontenc}
\usepackage{wrapfig,lipsum,booktabs}

\usepackage{soul}
\usepackage{dsfont}
\usepackage{enumerate}
\usepackage{enumitem}

\usepackage{amsmath}
\usepackage{amsfonts}
\usepackage{bbm}
\usepackage{dsfont}
\usepackage[Symbol]{upgreek}
\usepackage{lscape}
\usepackage{caption}
\usepackage{balance}
\usepackage{xspace}
\usepackage{float}

\usepackage{wasysym}
\usepackage[table,xcdraw,dvipsnames]{xcolor}
\usepackage{multirow}
\usepackage{array, boldline, rotating}

\usepackage{amssymb}
\usepackage{pifont}

\newcommand{\mc}[1]{\mathcal{#1}}

\renewcommand*\eqref[1]{(\ref{#1})}

\newcommand{\eg}{\emph{e.g.,~}}
\newcommand{\ie}{\emph{i.e.,~}}

\newcommand{\myparagraph}[1]{\vspace{0.07cm}\noindent\textbf{#1}~}

\def\code#1{\texttt{#1}}
\DeclarePairedDelimiter\norm{\lVert}{\rVert}

\newcommand{\thickhline}{\hlineB{4}}

\NewDocumentCommand{\supptitle}{s}{
\onecolumn
\begin{center}
    \rule{\textwidth}{0.03cm}\\[0.1cm]
    -Supplementary Material-\\[0.2cm]
    {\Large 
        \textbf{\mytitle }
    }\\
    \rule{\textwidth}{0.03cm}\\[0.2cm]
\end{center}
}

\definecolor{blue1}{rgb}{0.878431373,	0.921568627,	0.968627451}
\definecolor{blue2}{rgb}{0.654901961,	0.788235294,	0.870588235}
\definecolor{blue3}{rgb}{0.28627451,	0.505882353,	0.721568627}

\definecolor{LightCyan}{rgb}{0.88,1,1}
\definecolor{Blue}{rgb}{0, 0.5, 1}
\definecolor{Green}{rgb}{0.0, 0.8, 0.0 }
\definecolor{Red}{rgb}{0.95, 0.55, 0.6}
\definecolor{Skyblue}{rgb}{0.6, 0.6, 0.95 }
\definecolor{LightGray}{gray}{0.9}



\usepackage{framed}
\definecolor{shadecolor}{named}{LightGray}

\usepackage{PRIMEarxiv}

\usepackage[utf8]{inputenc} 
\usepackage[T1]{fontenc}    
\usepackage{hyperref}       
\usepackage{url}            
\usepackage{booktabs}       
\usepackage{amsfonts}       
\usepackage{nicefrac}       
\usepackage{microtype}      
\usepackage{lipsum}
\usepackage{fancyhdr}       
\usepackage{graphicx}       
\usepackage{times}
\usepackage{latexsym}
\usepackage{mathtools}

\usepackage{multirow}
\usepackage{adjustbox}
\usepackage{wrapfig}
\usepackage{kotex}
\usepackage{enumitem}
\usepackage{amsmath}
\usepackage{lipsum}
\usepackage{blindtext}

\usepackage{amsthm}
\usepackage{dsfont}
\usepackage{xcolor}
\usepackage{subcaption}

\hypersetup{
  colorlinks   = true, 
  urlcolor     = blue, 
  linkcolor    = MidnightBlue, 
  citecolor   = cyan 
}
\newcommand{\renewtheorem}[1]{%
  \expandafter\let\csname #1\endcsname\relax
  \expandafter\let\csname c@#1\endcsname\relax
  \expandafter\let\csname end#1\endcsname\relax
  \newtheorem{#1}%
}

\theoremstyle{plain}

\theoremstyle{definition}

\theoremstyle{theorem}
\renewtheorem{obs}[thm]{$\triangleright$ Observation}

\usepackage[capitalize,noabbrev]{cleveref}
\crefname{section}{Sec.}{Secs.}
\Crefname{section}{Section}{Sections}
\Crefname{table}{Table}{Tables}
\crefname{table}{Tab.}{Tabs.}

\def\equationautorefname~#1\null{Eq.~(#1)\null}

\title{\mytitle}

\author{
  Siqi Liang\textsuperscript{$*$} \\
  Michigan State University \\
  East Lansing, Michigan, USA \\
  \texttt{liangsi4@msu.edu} \\
   \And
  Sumyeong Ahn\thanks{Equal contribution.} \\
  Michigan State University \\
  East Lansing, Michigan, USA \\
  \texttt{sumyeong@msu.edu} \\
   \And
  Jiayu Zhou\thanks{Corresponding author.} \\
  Michigan State University \\
  East Lansing, Michigan, USA \\
  \texttt{jiayuz@msu.edu} \\
}

\begin{document}
\maketitle
\begin{abstract}
    Advancements in large language models (LLMs) have shown their effectiveness in multiple complicated natural language reasoning tasks. 
    A key challenge remains in adapting these models efficiently to new or unfamiliar tasks. 
    In-context learning (ICL) provides a promising solution for few-shot adaptation by retrieving a set of data points relevant to a query, called in-context examples (ICE), from a 
    training dataset and providing them during the inference as context. 
    Most existing studies utilize a centralized 
    training dataset, yet many real-world datasets may be distributed among multiple clients, and remote data retrieval can be associated with costs. 
    Especially when the client data are non-identical independent distributions (non-IID), retrieving from clients a proper set of ICEs needed for a test query presents critical challenges. 
    In this paper, we first show that in this challenging setting, test queries will have different preferences among clients because of non-IIDness, and equal contribution often leads to suboptimal performance.
    We then introduce a novel approach to tackle the distributed non-IID ICL problem when a data usage budget is present.
    The principle is that each client's proper contribution (budget) should be designed according to the preference of each query for that client.
    Our approach uses a data-driven manner to allocate a budget for each client, tailored to each test query. 
    Through extensive empirical studies on diverse datasets, our framework demonstrates superior performance relative to competing baselines. 
\end{abstract}
\section{Introduction}
\label{sec:intro}

Recent significant progress in large language models (LLMs)~\cite{achiam2023gpt, touvron2023llama, touvron2023llama2, team2023gemini} has demonstrated their effectiveness across various natural language processing (NLP) tasks~\cite{wang2018glue,wang2019superglue}. 
Despite their impressive performances, they still require adaptation to the specific downstream tasks for better performance.
However, adaptation poses challenges due to LLMs' vast number of trainable parameters. 

\begin{figure}[t]
    \centering
    \includegraphics[width=0.4\columnwidth]{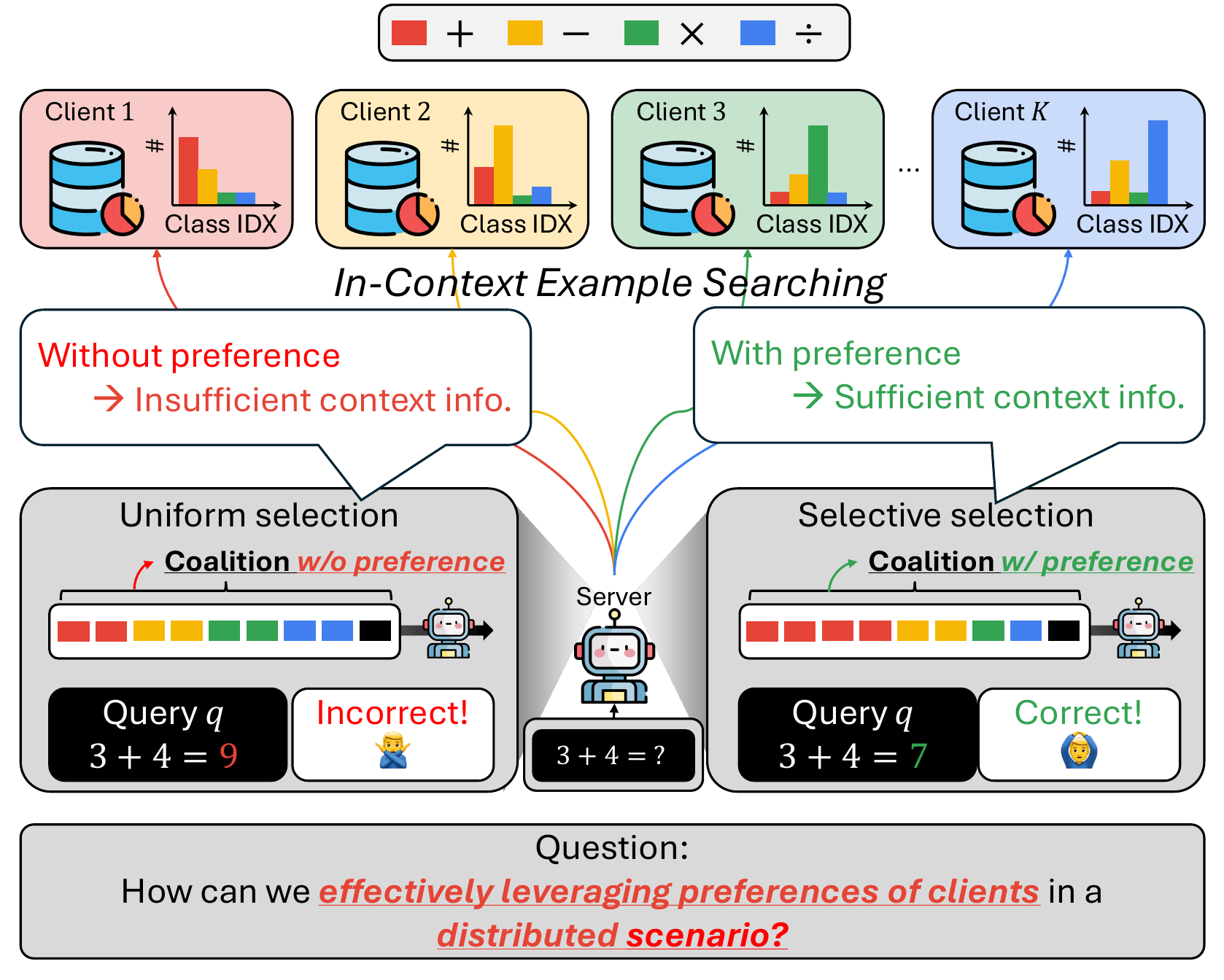}  
    \caption{Problem overview. When datasets are distributed among clients in a non-IID manner, it creates an obstacle in generating a good context (\textcolor{red}{left}). 
    However, by assigning appropriate budgets to leverage per-client expertise, better context can be created (\textcolor{ForestGreen}{right}).}
    \label{fig:intro}
\end{figure}

In-context learning (ICL)~\cite{dong2022survey} is a notable method that distinguishes itself through both its effectiveness and efficiency.
In brief, ICL adapts to the target task by incorporating context information following two primary steps: i) identify samples from the training dataset helpful to solve the target query by creating a prompt describing a context; ii) feed the constructed prompt with the target query and get the answer. 
Previous related works on ICL mainly have focused on the construction of a prompt describing the context, which involves several sub-problems, such as the retrieval of in-context examples (ICEs)~\cite{robertson2009probabilistic} and determining the optimal sequence for the selected ICEs~\cite{zhang2024batch}.



A common assumption in most existing ICL research is that the system has access to a high-quality centralized dataset used for retrieval. 
However, in many application scenarios, such as health informatics, centralized datasets may not be feasible, and data could be distributed in different institutions, which calls for the distributed ICL. 
In addition, when the data is proprietary and possesses high value towards inferences, access to data entries may also be bound to data pricing strategies~\cite{xu2023data,cong2022data}.
For instance, the system needs to pay the local institution based on the number of samples sent to the system \cite{tang2020abrief} as a means to share profits from inferences. 
Under this scenario, aggregating ICEs from local clients to a center server for ICL entails significant financial costs and lacks efficiency. 

In this paper, we focus on integrating knowledge from distributed clients to achieve better ICL performance under the per-query ICE budget constraint. Specifically, we formalize the distributed ICL problem where the ICEs are distributed on clients, and the server has an LLM for ICL inference but can only request a limited number of ICEs from all clients for each query, which we refer to as the \textit{ICE budget}.

We begin by identifying the key challenge in distributed ICL with ICE budget constraints lies in the non-independently and identically distributed (non-IID) training data, as shown in \autoref{sec:noniid-performance-drop}. 
For example, in~\autoref{fig:intro}, data samples are spread across $C$ clients, each with a unique data distribution. 
Specifically, client $1$ primarily contains \textcolor{red}{($+$)} samples, while client $2$ is mainly constituted by \textcolor{orange}{($-$)} examples. 
Only limited research~\cite{mohtashami2023social} tried to address the challenge of distributed datasets for ICL, while none considers the challenging real-world setting of non-IID clients. 
This leaves a critical question unanswered: \textit{What happens to distributed ICL when local clients are non-IID?}

To further the understanding of the key challenge in \textbf{the distributed non-IID ICL}, we explore the local retrieval process on non-IID clients. 
We found that each query has different preferences for different clients based on local knowledge distribution, that is, the number of samples needed from different clients should vary based on local sample distribution.
As the toy example shown in~\autoref{fig:intro}, when the server creates context by uniformly assigning budgets to clients, the answer might be incorrect due to the insufficiency of \textcolor{red}{($+$)} information in the context. 
To be more detailed, the server assigns the clients who have expertise on \textcolor{orange}{($-$)}, \textcolor{ForestGreen}{($\times$)}, and \textcolor{blue}{($\div$)} operations with the same budget as on \textcolor{red}{($+$)}, without any preference. 
Nevertheless, if the server assigns more budget to clients with many \textcolor{red}{($+$)} samples, such as client $1$, it can create a more relevant context to answer the query related to \textcolor{red}{($+$)} operation. 
This indicates that under non-IID, the server should allocate the budgets over clients based on the preference of each query itself, as well as the distribution of local training samples.

Motivated by this, we propose a novel distributed ICL framework to collaboratively collect scattered information among non-IID clients by properly assigning ICE budgets to each client. 
First, the server will gather the optimal budget statistics using an existing proxy dataset on the server side. 
Next, the server will use this dataset to train the budget allocator. 
During the deployment stage, the server will predict the proper budget for each client using this trained budget allocator given each test query and perform ICL among clients. 
Furthermore, in practical scenarios where privacy concerns arise, we augment our framework with the paraphrasing method~\cite{mohtashami2023social} to secure privacy.

\myparagraph{Contributions.} 
A summary of our contributions:
\begin{itemize}[leftmargin=10pt]
    \item To the best of our knowledge, we are the first to study the challenging real-world setting of ICL with distributed non-IID clients. 
    We identify the principal challenge as properly assigning the ICE budget for non-IID clients based on the preference of each test query and local knowledge distribution.
    \item We propose a framework to handle the distributed non-IID ICL. This framework trains a budget allocator on the server with the help of a server-side proxy dataset. Then, the server will use this trained allocator to decide how many ICEs to retrieve from each client for the ICL process, enabling collaborative action among clients.
    \item Across a range of dataset benchmarks featuring various non-IID configurations as well as on different LLM architectures, our approach has been validated to enhance ICL performance. 
    Notably, we examine both non-private, \ie communicate raw samples directly, and private cases using the paraphrasing method to secure privacy. In both scenarios, our approach shows superiority to the previous method and other reasonable baselines.
\end{itemize}

\section{Problem Formulation}
\label{sec:problem}
In this section, we provide a detailed problem formulation. First, we begin with the specifics of in-context learning (ICL), followed by a description of distributed non-IID ICL.

\begin{figure*}[t]
    \centering
    \includegraphics[width=0.6\textwidth]{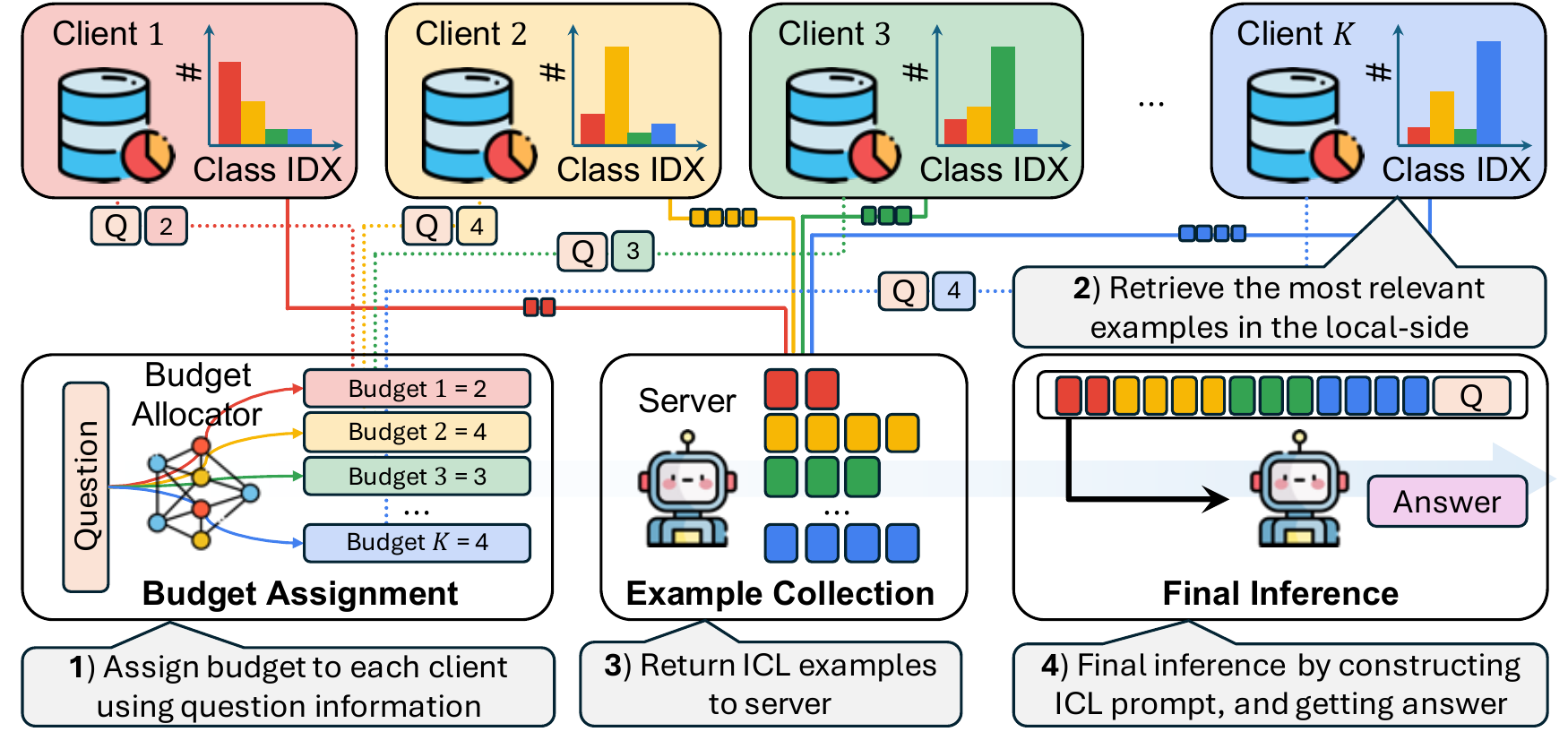}     
    \caption{Overview of the pipeline: First, the \textbf{budget allocator} assigns a budget to each client based on the question. Subsequently, each client retrieves their relevant samples and sends them back to the server. The server infers the answer by feeding the question, which is composed of concatenated context examples and the query.}
    \vspace{-5pt}
    \label{fig:frame}
\end{figure*}

\subsection{In-Context Learning}

\myparagraph{Notation.}
We consider a NLP tasks which have training dataset $\mc{D} = \{(x_{i}, y_{i})\}_{i=1}^{N}$ with $N$ training samples. 
Here, $x_i$ is the input text, and $y_i$ is the corresponding output. In the test phase, a test query $x_q$ is given.

\myparagraph{Retrieval.}
We employ the off-the-shelf pre-trained retriever  KATE~\cite{liu2021makes}\footnote{We do not fine-tune the retriever for each task, which is impractical because we cannot gather the distributed datasets.}, which utilizes $k$-NN example selection. This retriever employs a sentence encoder $\mc{E}(\cdot)$ to measure the similarity between the in-context example $x_{i}$ in dataset $\mc{D}$ and the query $x_q$ as follows:
\begin{align}\label{eq:score}
        d(e_i,e_q) = \norm{e_q - e_{i}}_2,
\end{align}
where $e_q = \mc{E}(x_{q})$ and $e_{i} = \mc{E}(x_{i})$.
We select $k$ samples using the following criterion:
\begin{equation}
\label{eq:topk}
    \mc{T}(e_q, k|\mc{D}) = \underaccent{e_i = \mc{E}(x_i) \forall {(x_{i}, y_{i}) \in \mc{D}}}{\arg \text{Top-}k} ( d(e_i,e_q) ),
\end{equation}
where $\mc{T}(e_q, k|\mc{D})$ denotes the selected samples from the dataset $\mc{D}$, and used for inference. 

\myparagraph{ICL Inference.} 
In the test phase, given a test query with input $x_i$, relevant $k$ training samples called in-context examples (ICEs) are selected, \ie $S = \mc{T}(e_q, k|\mc{D})$.
Based on the retrieved samples, we feed the constructed context prompt $s(S,x_q)$ into \code{LLM} for inference and obtain results via:
\begin{align}
\label{eq:inference}
    \begin{split}
            y_{t} &= \argmax_{y} p_{_\text{LLM}}( y | s(S,x_q), y_{<t} ) \\
           &s(S,x_q) = (x_1,y_1)\odot\ldots\odot(x_k,y_k)\odot x_q,
    \end{split}
\end{align}
where the $\odot$ operation denotes concatenation, and $s(S,x_q)$ is the context constructed using query $x_q$ and samples in $S$; 
the term $p_{_\text{LLM}}$ represents the output softmax probability of the LLM, functioning autoregressive, meaning that the output up to time $t$, \ie $y_{<t}$, is input back into the model to generate the $t^{\text{th}}$ output, $y_{t}$.
Previous works~\cite{ye2023compositional,levy2022diverse} on ICL mainly focus on the selection of $S$ under a centralized setting.
However, we investigate the scenario where $\mc{D}$ is split among several clients, each following non-IID distributions. 

\subsection{Distributed non-IID ICL}
\myparagraph{Distributed ICL Setting.}
We consider $C$ clients with a centralized server in our system. Each client $c \in [C]$ has local training dataset $\mc{D}_{c} = \{(x_{i}^{c}, y_{i}^{c})\}_{i=1}^{N_c}$ with $N_c$ training samples. 
Note that $\mc{D}_{c}$ follows different distributions for different clients. We follow the non-IID conditions as defined in~\cite{li2022federated}, with details provided in Appendix A.
In summary, we allocate data on a per-class basis, where each client receives a specific number of classes, meaning each client has samples from only specified classes.
Clients and the server have identical off-the-shelf pre-trained retrievers. Consider the computation resource limitation on clients as in many real scenarios~\cite{yoo2022open}, only the server is equipped with an \code{LLM}. Moreover, the server has limited proxy dataset $\mc{D}_{\text{proxy}} = \{(x_{j}^{\text{proxy}}, y_{j}^{\text{proxy}})\}_{j=1}^{N_{\text{proxy}}}$, that $N_{\text{proxy}} \ll \sum_{c=1}^C {N_c}$. 
The server has quite a small $\mc{D}_{\text{proxy}}$, and it is an auxiliary dataset to extract information for collaboration to make the problem feasible.

\myparagraph{Pipeline.}
First, the server requests relevant samples from each client by sending $x_q$ to all clients with local budgets $k_c$. 
Remark that each query $x_q$ has its own preference of each client, which can be represented as $k_c$.
Here, $x_q$ can be anonymized by paraphrasing, as done in previous works~\cite{mohtashami2023social}\footnote{Although our main experiments utilize the non-paraphrased dataset, we also present the paraphrased results in~\autoref{sec:exp}.}. 
Each client then selects the most relevant $k_c$ samples from their local training dataset, \ie $S_c = \mc{T}(e_q, k_c|\mc{D}_{c}) \subset \mc{D}_{c}$, and returns them to the server. 
The server receives $S_{c}$ from clients and generates the context $s$ based on the merged examples, $S = \bigcup_{c=1}^{C} S_c$. 
In the final step, the server infers $y$ using $s(S,x_q)$. The entire framework also can be described in~\autoref{fig:frame}. 
In this paper, we are concentrating on assigning $k_c$ to each client as described in~\autoref{fig:frame}.

\begin{figure}
    \centering
    \includegraphics[width=0.4\columnwidth]{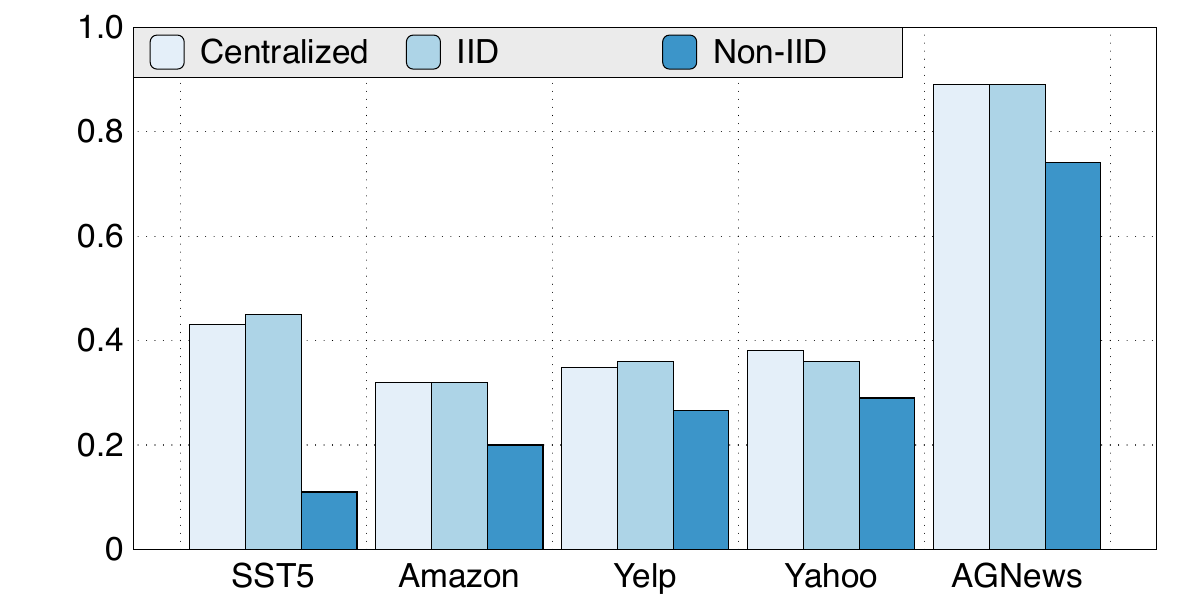}    
    \caption{Non-IID experimental results. It shows that centralized performance is comparable to the IID case, whereas non-IID scenarios exhibit a significant declined performance. This highlights the critical importance of addressing non-IIDness to find a solution.}
    \label{fig:noniid-motivation}
\end{figure}

\section{Observations}
\label{sec:motivation}
In this section, we describe several empirical supports to handle the distributed non-IID ICL. First, we demonstrate that non-IID distributions hinder the merging of scattered information. 
We then establish our goal, termed as \textbf{\emph{oracle budget}}, which reflects the server's preference for each client if the server knows all distributed data.
Finally, we check if predicting the oracle budget of each test query for inference is feasible.

\subsection{Non-IIDness Leads to Performance Drop}
\label{sec:noniid-performance-drop}
First of all, we evaluate the effect of non-IIDness. Straightforwardly, we distribute the budget $\{ k_c \}_{c=1}^C$ uniformly according to the following criteria: Given $C$ clients are involved in answering this question, and the number of samples for context is $k$. We first explore the na\"ive equally assigned local budget scheme in both IID and non-IID settings. That is, each client $c \in [C]$ locally retrieves top-$k_{c}$ samples where $k_{c} = \lceil \frac{k}{C} \rceil$ from local dataset $\mc{D}_{c}$. Detailed experimental settings are described in Appendix B.


As illustrated in~\autoref{fig:noniid-motivation}, we observe the followings: (1) There is no significant performance degradation between the centralized case (\textcolor{blue1}{$\blacksquare$}) and the IID case (\textcolor{blue2}{$\blacksquare$}). 
This is expected, as the merged top-$k_c$ samples in the IID case closely resemble the centralized top-$k$ samples. Any minor discrepancies are attributed to differences in sample ordering. (2) However, performance degradation becomes pronounced in non-IIDness case (refer to the comparison between \textcolor{blue3}{$\blacksquare$}, \textcolor{blue2}{$\blacksquare$} and \textcolor{blue1}{$\blacksquare$}). Hereinafter, we gather insights to address the distributed non-IIDness ICL.

\begin{figure}[t]
    \centering
    \begin{subfigure}[b]{0.235\textwidth}
        \includegraphics[width=\textwidth]{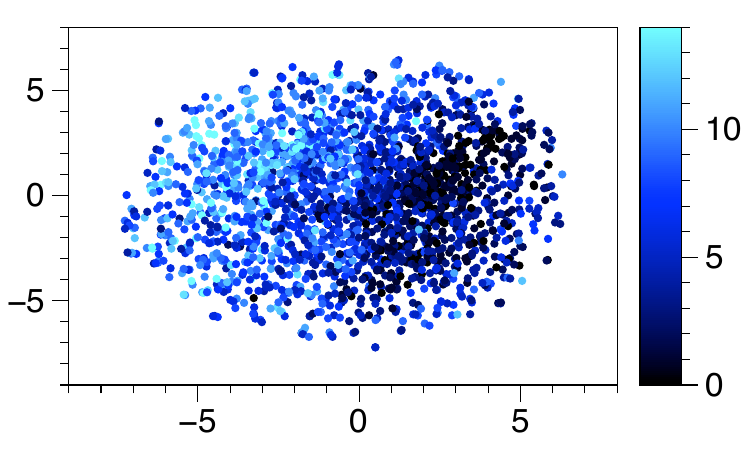}
        \includegraphics[width=\textwidth]{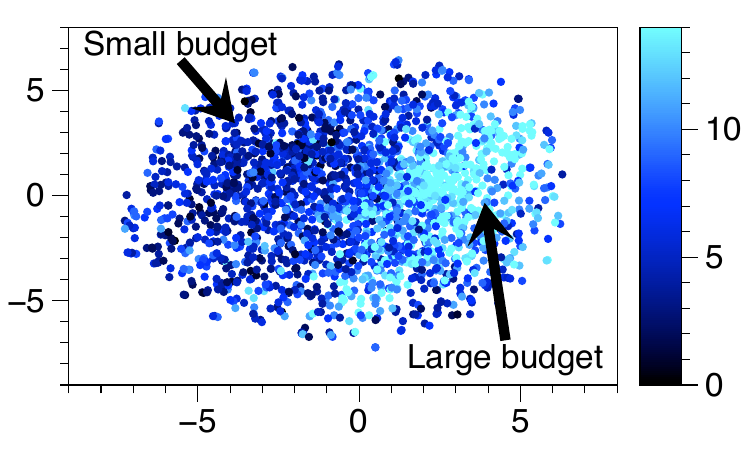}
        \caption{SST-5}
        \label{fig:tsne_sst5_c1}
    \end{subfigure}
    \begin{subfigure}[b]{0.235\textwidth}
        \includegraphics[width=\textwidth]{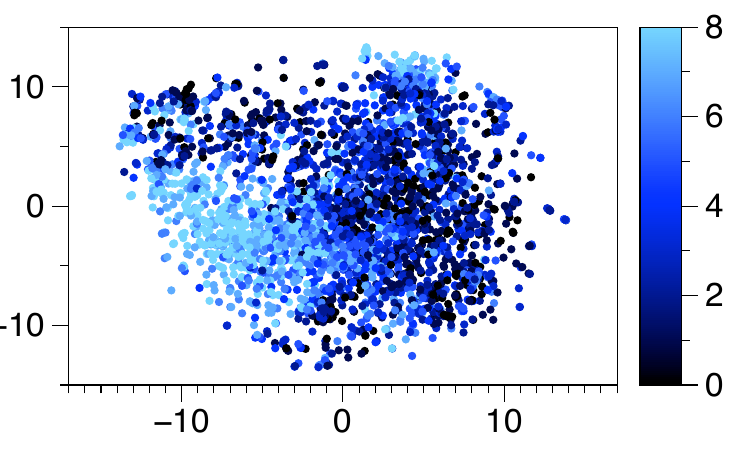}
        \includegraphics[width=\textwidth]{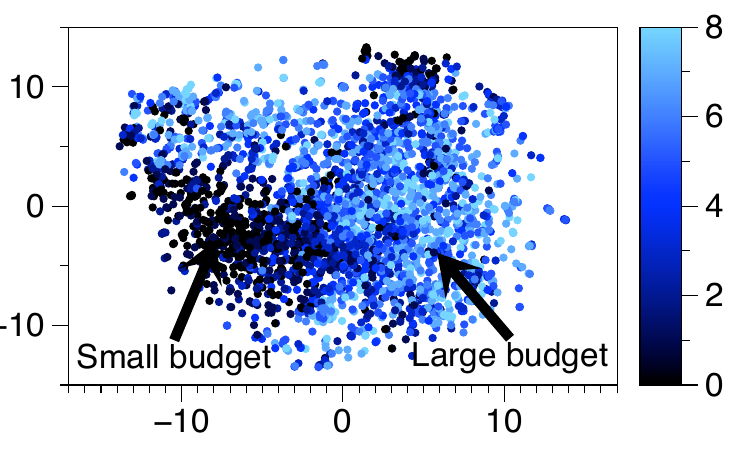}
        \caption{Yelp}
        \label{fig:tsne_yelp_c1}
    \end{subfigure}
    
    \caption{t-SNE analysis of each client across two datasets. The top and bottom rows depict oracle budget of client 1 and client 2, respectively. Each figure demonstrates that the budgets can be segregated by training a simple classifier, as they exhibit clustered subgroups.}
    \label{fig:tsne}
\end{figure}

\subsection{Proper Budget per Query for Each Client}
\label{sec:budget_assigning}
\myparagraph{Oracle budget.}
The remaining issue is that to make the server operate similar with the centralized manner, it needs to allocate the budget as if it knows complete knowledge of all clients. We call this budget for each client as the \emph{oracle budget} for query embedding $e_q$ and define it as follows:
\begin{equation*}
    k_c^\star(e_q) = \Big|\mc{T}(e_q, k|\mc{D}_c) \cap \mc{T}(e_q, k|\mc{D})\Big|,
\end{equation*}
where $\mc{T}(\cdot)$ is defined as~\autoref{eq:topk} and $|\cdot|$ is set cardinality. Note that the physical meaning of $k_{c}^{\star}(e_q)$ is the number of shared samples between the top-$k$ relevant to $e_q$ in local $\mc{D}_{c}$ and global $\mc{D}$ datasets.


\myparagraph{Check of predictability of oracle budget.}
For the next step, it is necessary to check if $e_q$ has sufficient patterns of oracle budget to extract and use it in the inference phase.
Our hypothesis is that similar queries may share similar oracle budget patterns and preferences on the same client, and it can lead to similar budget allocations for that client.
Therefore, to verify this hypothesis,
we perform t-SNE analysis~\cite{van2008visualizing} on the embeddings obtained from the retriever for queries. 
Furthermore, we color each sample based on the oracle budget $k_c^{\star}(e_q)$.
As described in~\autoref{fig:tsne}, similar query embeddings exhibit similar oracle budget patterns. 
This indicates that, given a test query, we can infer the budget assignment 
for each client. 
However, it is challenging to predict fine-grained budget value since there are no rigid classification patterns.
For instance, determining the detailed budget value seems challenging in the case of client $1$ in SST-5. 
Therefore, developing an efficient method to infer the exact 
budgets based on these broad patterns for each client are required.




\subsection{Observation Summary}
\label{sec:motivation-summary}

In summary, our findings and the approach for designing an algorithm are as follows: (1) non-IIDness significantly affects the distributed ICL setting, necessitating the development of a coalition method. To handle this problem, it is straightforward to allocate an appropriate number of budgets to each client, \ie making server work so as it knows client all samples. (2) By analyzing the query embeddings, we can determine the importance of each client per query.




\section{Method}
\label{sec:method}

\begin{figure}[t]
    \centering
    \includegraphics[width=0.4\columnwidth]{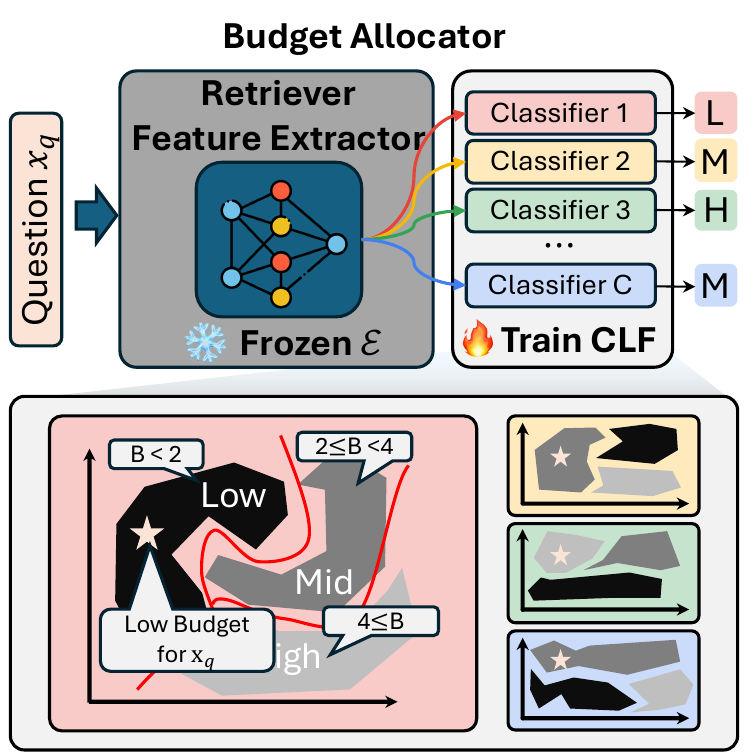}    
    \caption{Overview of the budget allocator: We train a budget allocator on top of the frozen feature extractor $\mc{E}$, which inherits from the retriever. During inference, when a test query $x_q$ is provided, this module determines the quantized budget levels for each client and allocates them accordingly.}
    \label{fig:budget_allocator}
\end{figure}

In this section, we outline the proposed algorithm to mitigate non-IIDness in the ICL framework. Specifically, we show how to train the \textbf{\emph{budget allocator}} and conduct inference. 

\begin{algorithm}[t]
\caption{Top-$k$ sampling, $\mc{T}(e,k| \mc{D})$}
\label{alg:client_top_score}
\begin{algorithmic}[1]
    \REQUIRE Query embedding $e$, Encoder $\mathcal{E}(\cdot)$ \\
    \textcolor{Skyblue}{{\footnotesize /* Compute embedding */ }}
    \FOR{$(x_i, y_i) \in \mc{D}$}
        \STATE $e_i \leftarrow \mathcal{E}(x)$ 
    \ENDFOR\\
    \textcolor{Skyblue}{{\footnotesize /* Select top-$k$ samples */ }}
    \STATE  $S = \underaccent{{(x_i,y_i) \in \mc{D}}}{\arg \text{Top-}k} \norm{e - e_i}_2$\\
    \RETURN $S$
\end{algorithmic}
\end{algorithm}

\begin{algorithm}[t]
\caption{Construct dataset}
\label{alg:construct}
\begin{algorithmic}[1]
    \REQUIRE Encoder $\mathcal{E}(\cdot)$, server-side ICE budget $k$, proxy dataset $\mathcal{D}_{\text{proxy}} = \{ (x_j, y_j) \}_{j=1}^{N_{\text{proxy}}}$ Quantization parameter $\delta$. \\
    \FOR{$(x_j^\text{proxy}, y_j^\text{proxy}) \in \mathcal{D}_{\text{proxy}}$}
        \STATE $e_j^\text{proxy} = \mathcal{E}(x_j^\text{proxy})$ \\
        \textcolor{Skyblue}{{\footnotesize /* Get distributed examples */ }}
        \FOR{$c \in [C]$} 
            \STATE $e_j^{\text{proxy}} \to \text{Client } c$ 
            \STATE $S_{c} = \mc{T}(e_j^{\text{proxy}}, k | \mc{D}_c)$
            \STATE Server $\leftarrow$ $S_c$
        \ENDFOR\\
        \textcolor{Skyblue}{{\footnotesize /* Construct optimal example */ }}
        \STATE $S = \bigcup_{c=1}^{C} S_c$
        \STATE $S^{\text{top}} = \underaccent{{(x_s,y_s) \in \mc{S}}}{\arg \text{Top-}k} \norm{e_j^{\text{proxy}} - \mc{E}(x_s)}_2$\\
        \textcolor{Skyblue}{{\footnotesize /* Compute proper budget size for each $c$ */ }}
        \STATE $k_{c}(e_j) = |S^{\text{top}} \cap S_c| // \delta $ \quad $\forall c \in [C]$
    \ENDFOR
    \STATE $B_{\text{proxy}} = \{(e_j, \{k_{c}(e_j)\}_{c=1}^{C})\}_{j=1}^{N_{\text{proxy}}}$
\RETURN $B_{\text{proxy}}$
\end{algorithmic}
\end{algorithm}

\begin{algorithm}[t]
\caption{Inference (\textcolor{Gray}{Client}, \textcolor{black}{Server})}
\label{alg:inference}
\begin{algorithmic}[1]
\REQUIRE Embedding model $\mathcal{E}(\cdot)$, LLM model $\mathcal{M}(\cdot)$, local datasets $\mathcal{D}_{c}$, budget allocator $f_c(\cdot)$ Buffering hyperparameter $\alpha$.
\ENSURE Test query $x_q$
\color{black}
\STATE Extract embedding $e_q = \mc{E}(x_q)$ \\
\color{black}
\FOR{$c \in [C]$}
    \color{black}
    \STATE $\hat{k}_c = f_c(e_q)$ 
    \STATE Send $e_q$ to all clients \\
    \color{Gray}
    \STATE $S_c = \mc{T}(e_q,\hat{k}_c + \alpha|\mc{D}_c)$ 
    \STATE return back $S_c \rightarrow$ Server
\color{black}
\ENDFOR
\color{black}
\STATE $S_{\text{agg}} = \bigcup_{c \in [C]} S_c$
\STATE $S = \mc{T}(e_q, k | S_{\text{agg}})$
\STATE $s(S,x_q) = (x_1,y_1)\odot...\odot(x_k,y_k)\odot x_q$ 
\RETURN $y = \mc{M}(s(S,x_q))$
\color{black}
\end{algorithmic}
\end{algorithm}
\subsection{Train a Budget Allocator}
\label{subsec:train_ba}

Based on~\autoref{sec:motivation}, it is feasible to assign budgets of each client by using the embeddings obtained from the retriever encoder $\mc{E}$. We first construct the datasets having the targeting budget values and then train the budget allocator.The pseudo-codes are described in~\autoref{alg:client_top_score}~and~\ref{alg:construct}.

\myparagraph{Construct dataset for oracle budget.}
First, we explain how to create a dataset to train the budget allocator for each client, as described in~\autoref{alg:construct}. 
Given proxy dataset $\mc{D}_{\text{proxy}}$, for all embeddings $e_j = \mc{E}(x_j)$ where $(x_j,y_j) \in \mc{D}_{\text{proxy}}$, we request $k$ samples from each client $c \in [C]$ using Top-$k$ procedure, \ie $S_c = \mc{T}(e,k|\mc{D}_c)$.
Once the server receives $k$ examples from each clients, \ie $\{ S_c \}_{c=1}^C$, it merges and re-orders them to obtains $S^{\text{top}}$.
Based on $S^{\text{top}}$, we count the number of samples from each client in $S^{\text{top}}$, \ie compute $k_{c}(e_j)$. After counting $k_{c}(e_j)$ for all clients, we quantize the budget levels for each client using the quantization hyper-parameter $\delta$. As a result, the output of this procedure is $B_{\text{proxy}}$ for all clients, composed of embeddings $e$ and their respective budgets $k_{c}(e_j)$.

\myparagraph{Train budget allocator.}
Based on the constructed dataset $B_{\text{proxy}}$, we train the \emph{budget allcoators}, \ie $\{f_c(\cdot)\}_{c=1}^C$, for each $f_c(\cdot)$ has Multi-layer perceptrons on top of the frozen feature extractor of the off-the-shelf retriever $\mc{E}$. The budget allcoators are trained on the cross-entropy loss, as we have already quantized the optimal budgets using the hyper-parameter $\delta$. Note that if $\delta$ is high, the quantization is severe, otherwise the quantization is mild. 

\subsection{Inference Using Budget Allocator}
\label{subsec:infer_ba}
We derive the response to the test query $x_q$ utilizing the \code{LLM} $\mc{M}(\cdot)$ through the described steps (see~\autoref{alg:inference} for specifics).
We first extract the embedding $e_q=\mc{E}(x_q)$. Then, we compute the allocated budget $\{\hat{k}_c=f_c(e_q)\}_{c=1}^{C}$ and send $\hat{k}_c$ to each client. Each client sends back top $\hat{k}_c + \alpha$ samples, \ie $S_c$, to the server. Note that we summarize how the budget allocator outputs $\hat{k}_c$ in~\autoref{fig:budget_allocator}. Here, $\alpha$ denotes the buffering hyper-parameter, which increases the chances for each client to be involved. After collecting $S_{\text{agg}} = \bigcup_{c \in [C]} S_c$, we aggregate them and run the usual ICL procedure.







\section{Experiment}
\label{sec:exp}

\subsection{Experiment setup}

First, we summarize the baselines, datasets, and the method for constructing non-IID settings. Finally, we depict the implementation details.

\myparagraph{Baselines.}
We compare our algorithm with various baselines, including social learning~\cite{mohtashami2023social}, which does not account for non-IIDness, and other possible ways for handling distributed non-IID ICL,
such as  Zero-shot, Proxy-only, Singleton (single client), Uniform-budget, Random-budget, and $\infty$-budget (oracle case). The detailed explanations are described in Appendix C.

\begin{table*}[t]
    \centering
    \resizebox{1.0\textwidth}{!}{
    \begin{tabular}{ccccccccccc}
    \thickhline
        \multicolumn{2}{c}{\multirow{2}{*}{Algorithm}} 
        & \multicolumn{7}{c}{Dataset}  &&\multirow{2}{*}{Avg}\\\cmidrule{3-9} 
        && SST-5 & Amazon & Yelp & MR & Yahoo & AGNews & Subj &&  \\\hline 
        Zero-shot 
                     && 29.19 & 24.70 
                     &  31.23 
                     &  73.95 & 25.87 
                     &  67.60 & 50.55 
                     && 43.30\\
                     \cmidrule{1-1} \cmidrule{3-9} \cmidrule{11-11}  
        Proxy-only
                     && 40.64{\small $\pm$ 2.89} & 28.43{\small $\pm$ 0.11} 
                     &  31.85{\small $\pm$ 1.28} 
                     &  70.40{\small $\pm$ 1.54} & 54.73{\small $\pm$ 0.93} 
                     &  84.65{\small $\pm$ 0.42} & 71.09{\small $\pm$ 1.34} 
                     && 54.54 \\ 
                     \cmidrule{1-1} \cmidrule{3-9} \cmidrule{11-11}  
        Singleton        
                     && 25.14{\small $\pm$ 4.18} & 24.03{\small $\pm$ 0.57} 
                     &  29.44{\small $\pm$ 3.91}  
                     &  50.00{\small $\pm$ 0.00}  & 38.14{\small $\pm$ 2.03}
                     &  50.60{\small $\pm$ 0.66}  & 50.00{\small $\pm$ 0.00} 
                     && 38.19 \\ 
                     \cmidrule{1-1} \cmidrule{3-9} \cmidrule{11-11} 
        Social Learning
                     && 36.03{\small $\pm$ 0.27} & 28.42{\small $\pm$ 0.19} 
                     &  29.25{\small $\pm$ 0.45} 
                     &  58.58{\small $\pm$ 0.18} & 46.03{\small $\pm$ 0.49}
                     &  81.10{\small $\pm$ 0.29} & 71.37{\small $\pm$ 0.71} 
                     && 50.11 \\ 
                     \cmidrule{1-1} \cmidrule{3-9} \cmidrule{11-11}  
        Uniform-budget
                     && 32.94 & 25.63
                     &  26.60 
                     &  33.65 & 43.00
                     &  73.17 & 63.20
                     && 42.60 \\ 
                     \cmidrule{1-1} \cmidrule{3-9} \cmidrule{11-11}  
        Random-budget
                     && 32.82{\small $\pm$ 0.82} & 25.69{\small $\pm$ 0.55} 
                     &  27.72{\small $\pm$ 0.51} 
                     &  34.68{\small $\pm$ 0.59} & 42.46{\small $\pm$ 0.53}
                     &  67.34{\small $\pm$ 0.39} & 65.37{\small $\pm$ 0.80} 
                     && 42.30 \\ 
                     \cmidrule{1-1} \cmidrule{3-9} \cmidrule{11-11}  
        \textcolor{gray}{$\infty$-budget}
                     && \textcolor{gray}{43.26} & \textcolor{gray}{32.70}
                     &  \textcolor{gray}{34.80} 
                     &  \textcolor{gray}{77.20} & \textcolor{gray}{62.67}
                     &  \textcolor{gray}{89.37} & \textcolor{gray}{91.4}                     
                     && \textcolor{gray}{61.62} \\
                     \cmidrule{1-1} \cmidrule{3-9} \cmidrule{11-11} 

        \textbf{Ours}
                     && \textbf{44.08{\small $\pm$ 0.12}} & \textbf{31.54{\small $\pm$ 0.22}} 
                     &  \textbf{35.48{\small $\pm$ 0.28}} 
                     &  \textbf{80.44{\small $\pm$ 0.67}} & \textbf{61.67{\small $\pm$ 0.25} }
                     &  \textbf{88.52{\small $\pm$ 0.30}} & \textbf{82.36{\small $\pm$ 0.91}} 
                     && \textbf{60.58} \\
    \thickhline
    \end{tabular}}
    \caption{Main results: To address the issue of non-IIDness in distributed ICL, we examined seven datasets and seven straightforward baselines. We run three random seeds and illustrate mean and std values. The top performance is highlighted in \textbf{bold} font, excluding the infinite budget scenario due to its impracticality. In summary, the proposed method effectively mitigates the non-iid distributed ICL problem to a reasonable extent.}
    \label{tab:main}
\end{table*}


\myparagraph{Datasets.}
We check the performance under $7$ datasets -- \underline{Sentiment classification}: SST-5~\cite{socher2013recursive}, Amazon~\cite{mcauley2013hidden}, Yelp~\cite{zhang2015character}, MR~\cite{pang2005seeing}, \underline{Topic classification}: Yahoo, AGNews~\cite{zhang2015character}, and \underline{Subjectivity classification}: Subj~\cite{pang-lee-2004-sentimental}. 

\myparagraph{Dataset partition for non-IIDness.}
We split the training dataset into $C$ subsets to ensure they follow a non-IID distribution. To achieve this, we partition the data based on class, following the splitting criteria outlined in \cite{li2022federated}. Specifically, each client has access to only $\gamma < \Gamma$ classes, where $\Gamma$ represents the total number of classes. We outline the summary of $\gamma$ for each dataset in Appendix D.

\myparagraph{Dataset paraphrasing.}
Due to concerns about sharing private samples between servers and clients, various techniques have been developed for natural language tasks. In this paper, we adopt the paraphrasing technique used in~\cite{mohtashami2023social}. Specifically, we utilize a small language model~\cite{team2024gemma}, designed for small terminal devices, to generate paraphrased questions. In Appendix E,
we summarize the instructions provided to the language model for rephrasing queries in the training dataset.

\myparagraph{Implementation details.}
We implement our method as well as baselines based on OpenICL~\cite{wu2023openicl}.
For the retriever scenario, we utilize the pre-trained \code{KATE} retriever~\cite{liu2021makes}, which has been trained on the SNLI~\cite{young-etal-2014-image} and MultiNLI~\cite{multinli} datasets. Note that they do not overlap with the datasets used in our experiment. They used RoBERTa-large~\cite{liu2019roberta} encoder model. We use \code{GPT-Neo-2.7B}~\cite{gpt-neo} pre-trained model as answering LLMs. hyper-parameters related to training budget allocators, $\alpha$, and $\delta$ are described in Appendix D
in detail.


\subsection{Main results}
We have presented the performance of our algorithm and baselines in~\autoref{tab:main}. 
First, we can observe that performance varies significantly depending on the way the budget is allocated, which indicates that the budget allocation scheme really matters in distributed non-IID ICL.
Additionally, even when using only the proxy dataset, there is a performance improvement, and this performance surpasses that of using other clients which have the tilted local datasets (\eg $29.19\% \to 40.64\%$ in SST-5 case). 
This indicates that utilizing a biased dataset can degrade the ICL performance. Although social learning algorithm has shown good performance in the previous paper, it does not perform well under the non-IID cases configured in this research. 
If we can use an infinite budget, all settings would exhibit high performance. 
However, our proposed algorithm demonstrates better performance than the infinite budget upper limit (\eg $34.86\% \to 35.48\%$ in the Yelp case). 
This is likely due to a mechanism that prevents unnecessary information from being selected by the retriever with high importance. 
Ultimately, the proposed algorithm shows an average performance improvement of $5.05\%$ percentage points across seven datasets compared to the best performance of baselines using the proxy dataset. 
This shows that the proposed algorithm can handle the non-IID case well.

\subsection{Analysis}
In this section, we further examine four key aspects: (1) privacy-preserving case analysis, which encompasses paraphrasing both training and testing queries, (2) sensitivity to hyper-parameters, (3) the performance of the trained budget allocator, and (4) the compatibility of the LLMs.

\begin{table*}[h]
    \centering
    \resizebox{0.5\columnwidth}{!}{
    \begin{tabular}{ccccccc}
    \thickhline
        \multicolumn{2}{c}{\multirow{2}{*}{Algorithm}} 
        & \multicolumn{3}{c}{Dataset}  &&\multirow{2}{*}{Avg}\\ \cmidrule{3-5} 
        && SST-5 & Yelp & Subj   &&  \\\hline 
        Zero-shot 
                     && 27.96 & 31.40 & 51.55 
                     && 36.97\\
                     \cmidrule{1-1} \cmidrule{3-5} \cmidrule{7-7}
        Proxy-only
                     && 39.39{\small $\pm$ 1.33} & 31.78{\small $\pm$ 1.75} & 73.46{\small $\pm$ 1.46} 
                     && 48.21\\
                     \cmidrule{1-1} \cmidrule{3-5} \cmidrule{7-7}
        Singleton        
                     && 25.31{\small $\pm$ 3.89} & 30.78{\small $\pm$ 4.88} & 50.08{\small $\pm$ 0.10} 
                     && 35.39\\
                     \cmidrule{1-1} \cmidrule{3-5} \cmidrule{7-7}
        Social Learning
                     && 33.09{\small $\pm$ 0.68} & 28.80{\small $\pm$ 0.33} & 74.82{\small $\pm$ 0.93} 
                     && 45.47\\
                     \cmidrule{1-1} \cmidrule{3-5} \cmidrule{7-7}
        Uniform-budget
                     && 27.06 & 26.60 & 63.30 
                     && 38.99\\
                     \cmidrule{1-1} \cmidrule{3-5} \cmidrule{7-7}
        Random-budget
                     && 27.29{\small $\pm$ 0.51} & 27.70{\small $\pm$ 0.46} & 63.88{\small $\pm$ 0.81} 
                     && 39.62\\
                     \cmidrule{1-1} \cmidrule{3-5} \cmidrule{7-7}
        \textcolor{gray}{$\infty$-budget}
                     && \textcolor{gray}{41.63} & \textcolor{gray}{37.23} & \textcolor{gray}{90.75} 
                     && \textcolor{gray}{56.54}\\
                     \cmidrule{1-1} \cmidrule{3-5} \cmidrule{7-7}

        Ours
                     && \textbf{40.37}{\small $\pm$ 0.27} & \textbf{36.52}{\small $\pm$ 0.89} & \textbf{83.82}{\small $\pm$ 1.00}
                     && \textbf{53.57}\\
        
    \thickhline
    \end{tabular}
    }
    \caption{Analysis of the generated query and training samples: We re-create the datasets using small-sized LLMs and conduct the experiments as in~\autoref{tab:main} under the exactly same experimental settings.}
    \label{tab:paraphrase}
\end{table*}
\myparagraph{Paraphrasing results.}
Due to privacy concerns in the fundamental distributed system, we evaluate the performance of paraphrased datasets, with results detailed in~\autoref{tab:paraphrase}. Our method demonstrates superior performance compared to other baselines across multiple datasets. We used the exact same data settings as in~\autoref{tab:main}. 
Specifically, performance on the Subj and SST-5 datasets is lower than without paraphrasing, while the Yelp dataset shows a slight improvement. 
Additionally, as consistent with~\autoref{tab:main}, non-IIDness causes significant performance degradation for ICL methods, as seen by comparing Zero-shot with ICL-related methods (\eg $27.96\% \to 25.31\%$ in the Singleton case).

\begin{figure}[h]
    \centering
    \begin{subfigure}[b]{0.3\columnwidth} 
        \includegraphics[width=0.95\textwidth]{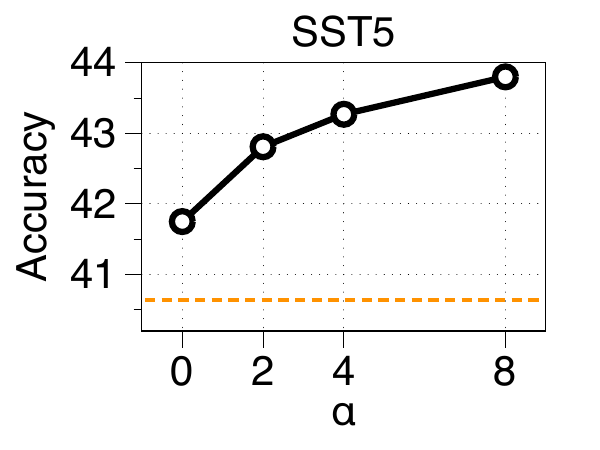}
        \vspace{-5pt}
        \includegraphics[width=0.95\textwidth]{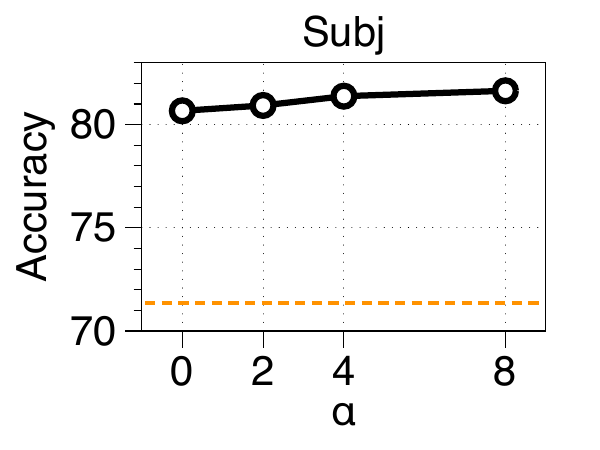}
        \vspace{-5pt}
        \caption{Additional budget $\alpha$ analysis \textcolor{blue}{[need to update]}}
        \label{fig:alpha}
    \end{subfigure}
    \begin{subfigure}[b]{0.3\columnwidth} 
        \includegraphics[width=0.95\textwidth]{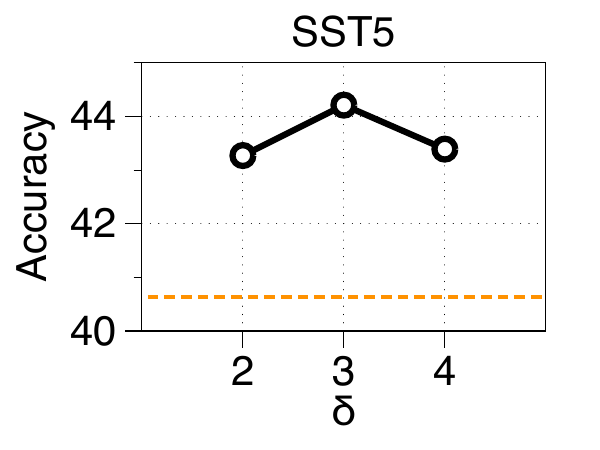}
        \vspace{-5pt}
        \includegraphics[width=0.95\textwidth]{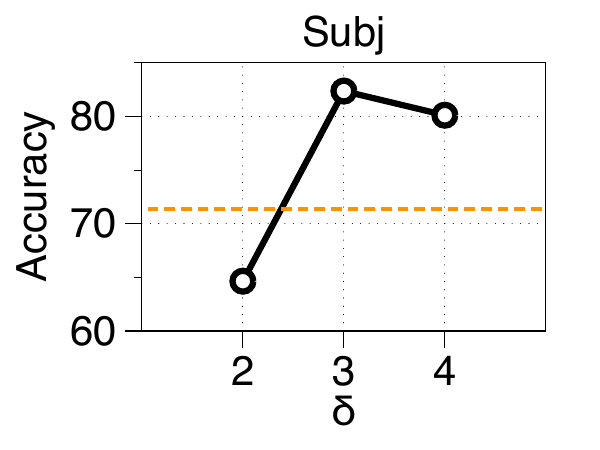}
        \vspace{-5pt}
        \caption{Resolution of budget allocator $\delta$ analysis \textcolor{blue}{[need to update]}}
        \label{fig:delta}
    \end{subfigure}    
    \caption{Hyperparameter, $\delta$ and $\alpha$, analysis. We check SST-5 and Subj datasets under \code{GPT-Neo-2.7B}. The \textcolor{orange}{orange line} indicates the 2nd best performance in~\autoref{tab:main}.}
    \label{fig:hparam_analysis}
\end{figure}

\begin{figure}[h]
    \centering
    \begin{subfigure}[b]{0.33\columnwidth} 
        \includegraphics[width=0.95\textwidth]{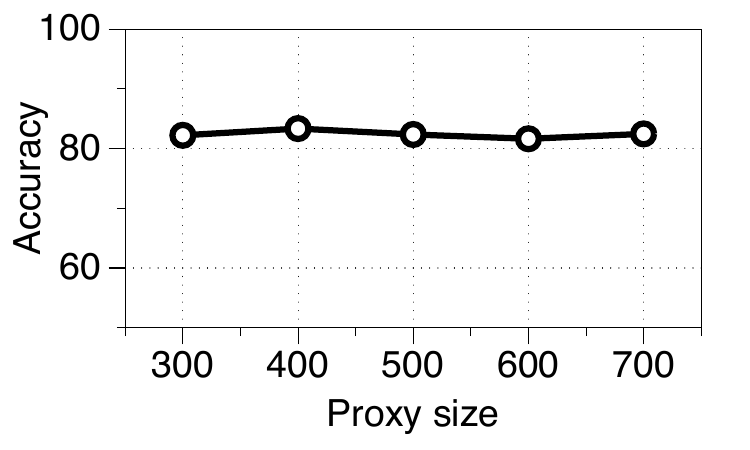}
    \end{subfigure}    
    \caption{Proxy size analysis. We check Subj dataset under \code{GPT-Neo-2.7B}.}
    \label{fig:proxy_size_analysis}
\end{figure}

\myparagraph{Hyper-parameter sensitivity.}
We examine the sensitivity of the hyper-parameters of our method. 
We have two hyper-parameters: $\delta$, which is the resolution of the budget allocator; $ \alpha$, which represents the additional budget allocated to each client as a buffer; and proxy size, which is the size of proxy data for the budget allocator training. 
As illustrated in~\autoref{fig:hparam_analysis}, when we increase $\alpha$, the performance is improved while the budget efficiency is reduced. 
On the other hand, when $\delta$ is high (or low), it has too dense (or sparse) representation of the budget class, thus performance is degraded. Nevertheless, the performance is higher than the other baselines in~\autoref{tab:main}. For the sensitivity of the size of proxy data, it is revealed that our framework is not sensitive to how many proxy data samples are used to train the budget allocator, as shown in~\autoref{fig:proxy_size_analysis}. This indicates our method is stable even with limited proxy data on the server side.

\begin{figure}[h]
    \centering
    \begin{subfigure}[b]{0.3\columnwidth} 
        \includegraphics[width=\textwidth]{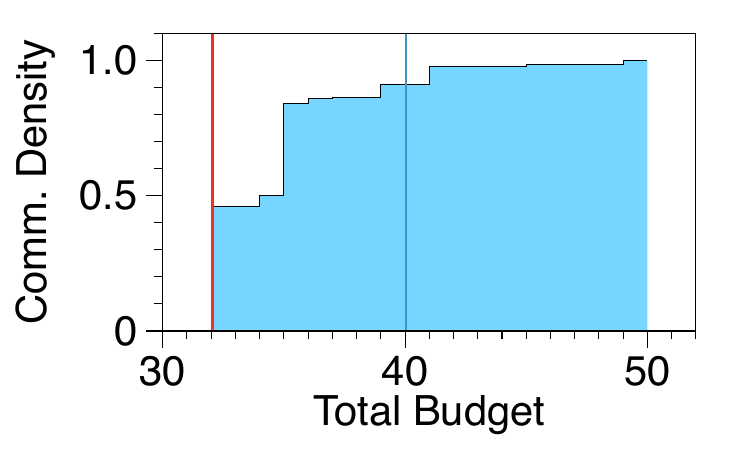}
        \caption{SST-5}
        \vspace{-5pt}
        \label{fig:budget_sst5}
    \end{subfigure}
    \begin{subfigure}[b]{0.3\columnwidth} 
        \includegraphics[width=\textwidth]{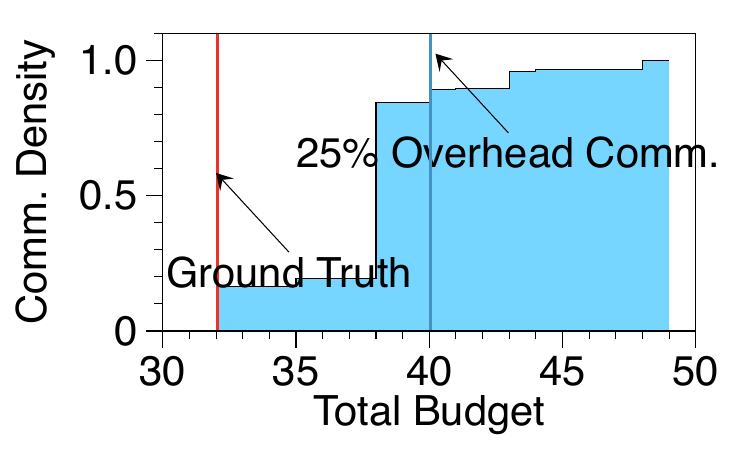}
        \caption{MR}
        \vspace{-5pt}
        \label{fig:budget_mr}
    \end{subfigure}
    \caption{Analyze the allocated budget. Analyze the total amount of budget allocated to clients under two datasets. \textcolor{red}{Red} and \textcolor{blue}{blue} lines denote the oracle and $25\%$ larger total budgets compared to the oracle case.}
    \label{fig:budget_analysis}
\end{figure}

\myparagraph{Trained budget allocator.}
We assess whether the trained budget allocator distributes budgets appropriately for each client.
To evaluate efficiency, we examine the number of samples, \ie $\hat{k}_{c}$ communicated for all queries and plot a histogram. 
As demonstrated in \autoref{fig:budget_analysis}, we confirm that the proposed algorithm's forecasts exhibit nearly identical performance to the oracle budget when an additional $25\%$ budget is allocated. 
Note that without the proposed algorithm, it is necessary to assign  $k \times C$ number of budgets to get a performance similar to the oracle case.

\begin{table*}[h]
    \centering
    \resizebox{0.55\columnwidth}{!}{
    \begin{tabular}{cccccc}
    \thickhline
        \multirow{2}{*}{Algorithm}
                            && \multicolumn{3}{c}{Architecture}  \\ \cmidrule{3-5}
                            && GPT-Neo-1.3B  & GPT-Neo-2.7B 
                            & Llama-2-7B   \\\hline 
        Zero-shot           && 51.30         
                             & 50.55        
                             & 49.10 \\
        Proxy-only          && 80.18{\small $\pm$ 1.87}         
                             & 71.09{\small $\pm$ 1.34}        
                             & 88.13{\small $\pm$ 0.74} \\
        Singleton           && 50.00{\small $\pm$ 0.00}        
                            & 50.00{\small $\pm$ 0.00}        
                            & 52.89{\small $\pm$ 3.43} \\
        Social Learning     && 68.55{\small $\pm$ 0.64}         
                            & 71.37{\small $\pm$ 0.71}        
                            & 88.82{\small $\pm$ 0.50} \\
        Uniform-budget      && 44.40         
                             & 63.20        
                             &   54.00 \\
        Random-budget       && 43.68{\small $\pm$ 0.80}         
                             & 65.37{\small $\pm$ 0.80}        
                             & 55.60{\small $\pm$ 0.41}\\
        \textcolor{gray}{$\infty$-budget}
                            && \textcolor{gray}{92.05}         
                            & \textcolor{gray}{91.40}        
                            & \textcolor{gray}{92.30}\\ \hline
        Ours                && \textbf{85.73}{\small $\pm$ 0.94}         
                             & \textbf{82.36{\small $\pm$ 0.91}}       
                             & \textbf{91.58}{\small $\pm$ 0.14} \\
    \thickhline
    \end{tabular}
    }
    \caption{Default non-IID setting of Subj using different LLMs. 32 ICEs for server LLM inference}
    \label{tab:architecture_type_subj}
\end{table*}
\myparagraph{Other types of LLMs.}
We utilize various LLM architectures to assess the compatibility of the proposed algorithm. Specifically, we evaluate the SST-5 dataset using different model sizes, including \code{GPT-Neo-1.3B}~\cite{gpt-neo} and the latest architecture, the \code{Gemma-2B}~\cite{team2024gemma} model. As demonstrated in \autoref{tab:architecture_type_subj}, the proposed algorithm exhibits a plug-and-play capability and achieves reasonable performance improvements in the distributed non-IID ICL setting.
\section{Related Work}
\label{sec:related}

\myparagraph{In-context learning.}
ICL~\cite{dong2022survey} is one of the fastest paradigms using pre-trained LLMs by feeding several examples to construct the context to solve the given query. 
The main criteria of this research field are to find the most informative samples among the training datasets. 
For example~\cite{liu2021makes} trains BERT~\cite{devlin2018bert} oriented encoder and use the $k$ nearest neighbors. 
One of the reasonable sparse retriever, rule-based approach, is using BM25~\cite{robertson2009probabilistic} which measures the term-frequency. 
\cite{rubin-etal-2022-learning} proposed an efficient retriever called EPR. 
It trains two encoders by inheriting the method of dense passage retriever (DPR)~\cite{karpukhin2020dense} under the loss of positive and negative pairs. 
To reduce the domain specificity, \cite{li2023unified} proposed UDR, which is applicable to multiple domain tasks in a universal way and shows reasonable performance from a single retriever. 
PromptPG~\cite{lu2022dynamic} utilized a reinforcement learning framework to train the retriever so that it can generate context to improve the answerability of LLMs. 
Similarly, LLM-R~\cite{wang2023learning} uses a reward model to train the retriever. 
Note that extensive research has not targeted to solve the distributed cases. These works have seen the centralized case.

\myparagraph{Distributed ICL.}
To the best of our knowledge, only a single study~\cite{mohtashami2023social} tries to address ICL in a distributed manner. 
However, this paper solely focuses on merging the distributed information without considering the nature of the non-identically distributed information. 
Many studies, such as those on federated learning~\cite{li2021survey, zhang2021survey, mammen2021federated}, address the non-IID distribution of datasets, highlighting the need to handle distributed non-IID ICL.

\section{Conclusion}
\label{sec:conclusion}
In this paper, we tackle the challenge of ICL when datasets are distributed among clients with non-IID distributions. 
Initially, we examine if non-IID distributions lead to performance degradation and discover that they cause significant drops in performance. 
We propose an algorithm that learns the task of budget assignment and employs it during inference to allocate appropriate budgets for each query. 
Using this proposed algorithm, we achieve performance improvements across various benchmarks. 


\clearpage    


\bibliographystyle{unsrt}  
\bibliography{ref}



\end{document}